\documentclass[10pt,twocolumn,letterpaper]{article}

\usepackage{cvpr}
\usepackage{times}
\usepackage{epsfig}
\usepackage{graphicx}
\usepackage{amsmath}
\usepackage{amssymb}

\usepackage[pagebackref=true,breaklinks=true,letterpaper=true,colorlinks,bookmarks=false]{hyperref}
\cvprfinalcopy % *** Uncomment this line for the final submission

\def\cvprPaperID{****} % *** Enter the CVPR Paper ID here

\begin{document}

%%%%%%%%% TITLE
\title{UAVid: A Semantic Segmentation Dataset for UAV Imagery}

\author{Ye Lyu \hspace{3mm} George Vosselman\\
\small University of Twente\\
\small The Netherlands\\
\and Gui-Song Xia\\
\small Wuhan University\\
\small China\\
\and Alper Yilmaz\\
\small Ohio State University\\
\small USA\\
\and Michael Ying Yang\\
\small University of Twente\\
\small The Netherlands}

%\and

\maketitle

\begin{abstract}
	Semantic segmentation has been one of the leading research interests in computer vision recently. It serves as a perception foundation for many fields, such as robotics and autonomous driving. The fast development of semantic segmentation attributes enormously to the large scale datasets, especially for the deep learning related methods.
	There already exist several semantic segmentation datasets for comparison among semantic segmentation methods in complex urban scenes, such as the Cityscapes and CamVid datasets, where the side views of the objects are captured with a camera mounted on the driving car. There also exist semantic labeling datasets for the airborne images and the satellite images, where the top views of the objects are captured. However, only a few datasets capture urban scenes from an oblique Unmanned Aerial Vehicle (UAV) perspective, where both of the top view and the side view of the objects can be observed, providing more information for object recognition.
	In this paper, we introduce our UAVid dataset, a new high-resolution UAV semantic segmentation dataset as a complement, which brings new challenges, including large scale variation, moving object recognition and temporal consistency preservation.
	Our UAV dataset consists of 30 video sequences capturing 4K high-resolution images in slanted views. In total, 300 images have been densely labeled with 8 classes for the semantic labeling task. We have provided several deep learning baseline methods with pre-training, among which the proposed Multi-Scale-Dilation net performs the best via multi-scale feature extraction, reaching a mean intersection-over-union (IoU) score around 50\% and outperforming the others by more than 1.6\%. We have also explored the influence of spatial-temporal regularization for sequence data by leveraging on feature space optimization (FSO) and 3D conditional random field (CRF), which improves the mean IoU scores by around another 0.5\%. Our UAVid website and the labeling tool have been published~\footnote{\url{https://uavid.nl/}}.
\end{abstract}

\section{Introduction}
Visual scene understanding has been advancing in recent years, which serves as a perception foundation for many fields such as robotics and autonomous driving. The most effective and successful methods for scene understanding tasks adopt deep learning as their cornerstone, as it can distill high-level semantic knowledge from the training data. However, the drawback is that deep learning requires a tremendous number of samples for training to make it learn useful knowledge instead of noise, especially for real-world applications.
Semantic segmentation, as part of scene understanding, is to assign labels for each pixel in the image. In order to make the best of deep learning methods, a large number of densely labeled images are required.

{At present, there are several public semantic segmentation datasets available, which focus only on common objects in natural images. They all capture images from the ground. The Ms-COCO~\cite{MSCoco} and the Pascal VOC~\cite{PascalVOC} datasets provide semantic segmentation tasks for common object recognition in common scenes. They focus on classes like person, car, bus, cow, dog, and other objects. In order to help semantic segmentation models generalize better across different scenes, the ADE20K dataset~\cite{ADE20k} spans more diverse scenes. Objects from much more different categories are labeled, bringing more variability and complexity for object recognition. The above datasets are often used for common object recognition.}

{There are more semantic segmentation datasets designed using street scenes for autonomous driving scenarios~\cite{CamVid,HighwayDriving,Cordts2016Cityscapes,Daimler_Urban_Segmentation,deepdrive,kitti}. Images are captured with cameras mounted on vehicles. The objects of interest include pedestrians, cars, roads, lanes, traffic lights, trees, and other surrounding objects near the streets. Especially, the CamVid~\cite{CamVid} and the Highway Driving~\cite{HighwayDriving} datasets provide continuously labeled driving frames, which can be used for video semantic segmentation with temporal consistency evaluation. The Cityscapes dataset~\cite{Cordts2016Cityscapes} focuses more on the data variation. It is larger in the number of images and the size of each image. Images are collected from 50 cities, making it closer to real-world complexity.}

{Regarding the remote sensing platforms, the number of datasets for semantic segmentation is much smaller, and the images are often captured in the nadir view, in which only the top of the objects can be seen.
For the airborne imagery, the ISPRS 2D semantic labeling benchmarks~\cite{ISPRSbenchmark14} provide Vaihingen and the Potsdam datasets targeting on semantic labeling for the urban scenes. There are $6$ classes defined for the semantic segmentation task, including impervious surfaces, building, low vegetation, tree, car ,and background clutter. The Vaihingen and the Potsdam datasets are $9$ cm and $5$ cm resolutions, respectively. Houston  dataset~\cite{HoustonCampus} provides hyperspectral images (HSIs) and Light Detection And Ranging (LiDAR) data, both of which have $2.5$m spatial resolutions, for the pixel level region classification. Zeebruges~\cite{zeebruges} provides a dataset with 7-tiles. There are $8$ classes defined for both the land cover and the object classification. Besides the same $6$ class types as in the ISPRS 2D semantic labeling datasets, additional boat and water classes are included. The RGB images are of $5$ cm resolutions.
For the satellite imagery, the DeepGlobe benchmarks~\cite{deepglobe} provide a semantic labeling dataset for the land cover classification with a pixel resolution of $50$ cm. The images are of the sub-meter resolution, covering $7$ classes, i.e., urban, agriculture, rangeland, forest, water, barren, and unknown. GID dataset~\cite{GID} offers $4$m resolution multispectral (MS) satellite images from Gaofen-2 (GF-2) imagery for the land use classification. The target classes include $15$ fine classes belonging to $5$ major categories, which are built-up, farmland, forest, meadow, and water.}

All the datasets above have had high impacts on the development of current state-of-the-art semantic segmentation methods. However, there are few high-resolution semantic segmentation datasets~\cite{aeroscapes} based on UAV imagery with slanted views, which is supplemented with our UAVid dataset.
The unmanned aerial vehicle (UAV) platform is more and more utilized for the earth observation. Compact and light-weighted UAVs are a trend for future data acquisition. The UAVs make image retrieval in large areas cheaper and more convenient, which allows quick access to useful information around a certain area. Distinguished from collecting images by satellites and airplanes, UAVs capture images from the sky with flexible flying schedules and higher spatial resolution, bringing the possibility to monitor and analyze landscape at specific locations and time swiftly. 

The inherently fundamental applications for UAVs are surveillance~\cite{UAV_Surveillance1,UAV_Surveillance2} and monitoring~\cite{UAV_Monitoring} in the target area. They have already been used for smart farming~\cite{smartfarm}, precision agriculture~\cite{Algriculture}, and weed monitoring~\cite{UAV_monitor_weed}, but few researches have been done for urban scene analysis. The semantic segmentation research for urban scenes could be the foundation for applications such as traffic monitoring, e.g., traffic jams and car accidents, population density monitoring and urban greenery monitoring, e.g., vegetation growth and damage. 
Although there are existing UAV datasets for detection, tracking, and behavior analysis~\cite{visdrone,uav_det,uav123,UAV_behavior}, to the best of our knowledge, there exists only one low altitude UAV semantic segmentation dataset before our UAVid, namely the Aeroscapes~\cite{aeroscapes} dataset. Our UAVid dataset is comprised of much larger images that capture scenes in much larger range and with more scene complexity regarding the number of objects and object configurations, which make our UAVid dataset better for UAV urban scene understanding than the Aeroscapes dataset.

In this paper, a new UAVid semantic segmentation dataset with high-resolution UAV images in slanted views has been brought out, which is designed for the semantic segmentation of urban scenes. We have brought out several challenges for the new dataset: the large scale variation between objects in different distances or of different categories, the moving object (separation of moving cars and static cars) recognition in the urban street scene and the preservation of the temporal consistency for better predictions across frames. These challenges mark the uniqueness of our dataset. {In total, 300 high-resolution images from 30 video sequences are labeled with 8 object classes. The size of our dataset is ten times of the Vaihingen dataset~\cite{ISPRSbenchmark14}, five times of the CamVid dataset~\cite{CamVid} and twice of the Potsdam dataset~\cite{ISPRSbenchmark14} regarding the labeled number of pixels.} All the labels are acquired with our in-house video labeler tool. Besides the provided image-label pairs, which are acquired with 0.2 FPS, unlabeled images are also provided with 20 FPS for users. The additional images are provided to aid the object recognition potentially. 
To provide performance references for the semantic labeling task and to test the usability of our dataset, several typical deep neural networks (DNNs) are utilized, including FCN-8s~\cite{FCN8s}, Dilation net~\cite{dilationNet} and U-Net~\cite{UNet}, which are widely used and stable for semantic segmentation task across different datasets. In addition, we propose a novel multi-scale-dilation net, which is useful to handle the problem of large scale variation that is prominent in the UAVid dataset. 
In order to benefit from preserving the consistent prediction across the frames, an existing spatial-temporal regularization method (FSO~\cite{FSO}) is applied for post-processing. All the DNNs combined with FSO are evaluated as baselines.

By bringing the urban scene semantic segmentation task for the UAV platform, researchers could gain more insights for the visual understanding task in the UAV scenes, which could be the main foundations for higher level smart applications. As the data from UAVs has its own specialties, semantic segmentation task using UAV data deserves more attention.

\begin{figure*}[ht]
\centering
\includegraphics[width=1.0\textwidth]{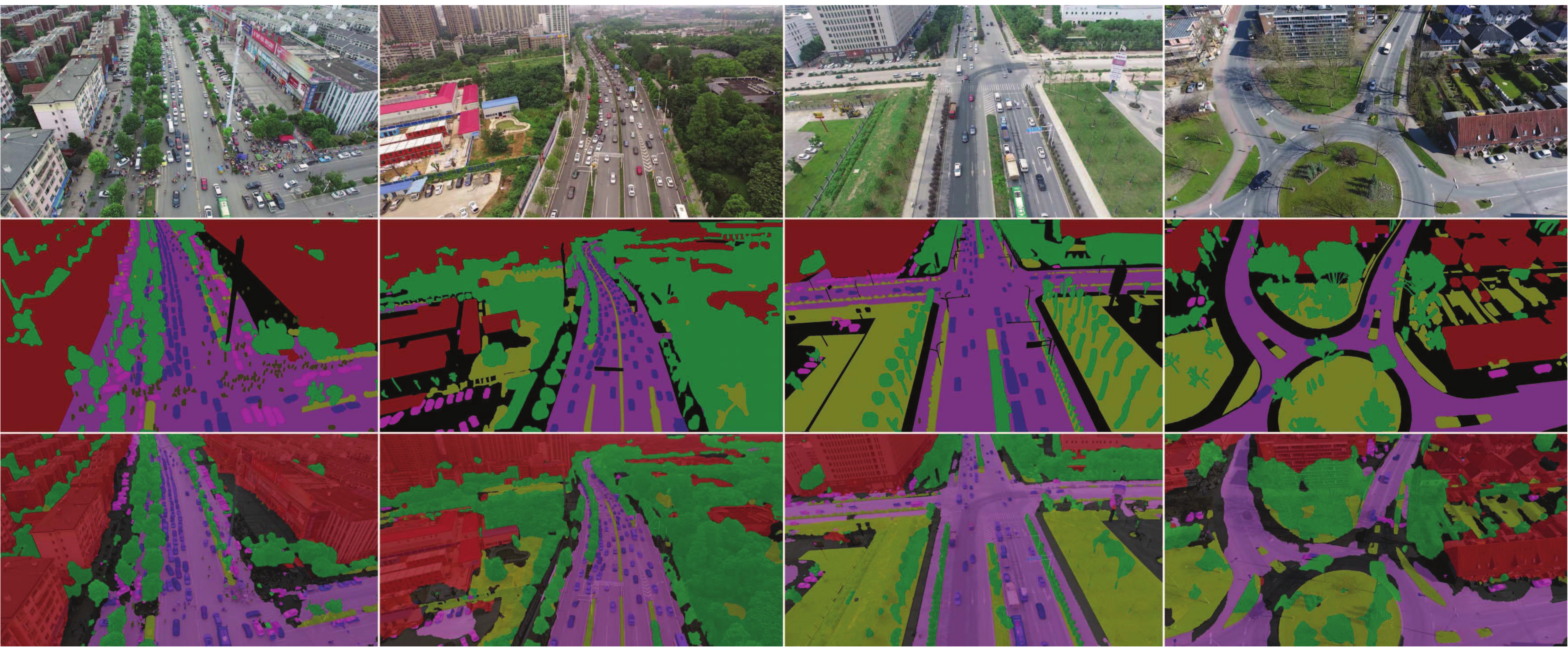}
\includegraphics[width=0.80\textwidth]{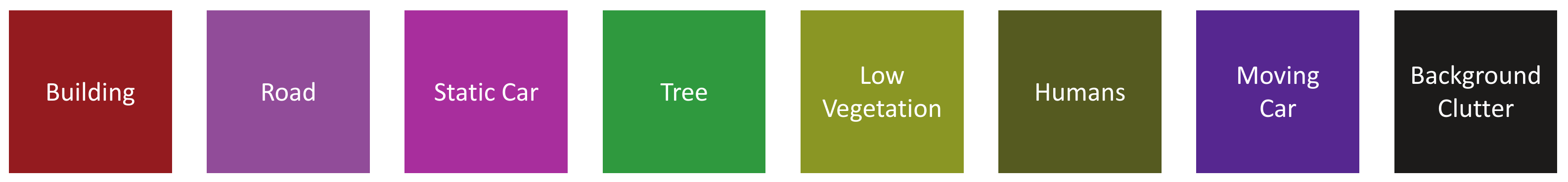}
%\textcolor{red}{uncomment for image}
%\includegraphics[scale=1.0]{figurefile}
\caption{\textbf{Example images and labels from UAVid dataset.} First row shows the images captured by UAV. Second row shows the corresponding ground truth labels. Third row shows the prediction results of MS-Dilation net+PRT+FSO model as in Tab.~\ref{tb_IoU}.}
\label{fig_dataset_eg}
\end{figure*}

The rest of the paper is organized as follows. Section~\ref{sec:ds} details how the UAVid dataset is built for the urban scene semantic segmentation, including the data specification, the class definition, the annotation methods, and the dataset splits. Section~\ref{sec:task} presents the semantic labeling task for the UAVid dataset. The section involves the task illustration and the baseline methods for the task. Section~\ref{sec:exp} shows the corresponding experiment results with the analysis for the baseline methods. Lastly, section~\ref{sec:conclusion} provides the concluding remarks and the prospects for the UAVid dataset.

\section{Dataset} ~\label{sec:ds}
Designing a UAV dataset requires careful thought about the data acquisition strategy, UAV flying protocol, and object class selection for annotation. The whole process is designed considering the usefulness and effectiveness for the UAV semantic segmentation research. In this section, the way to establish the dataset is illustrated. Section~\ref{sec:data_spec} shows the data acquisition strategy. Section~\ref{sec:class_def} and ~\ref{sec:anno} describe the classes for the task and the annotation methods respectively. Section~\ref{sec:ds_split} illustrates the data splits for the semantic segmentation task.

	\subsection{Data Specification} ~\label{sec:data_spec}
	Our data acquisition and annotation methodology is designed for UAV semantic segmentation in complex urban scenes, featuring on both static and moving object recognition.  
	In order to capture data that contributes the most towards researches on UAV scene understanding, the following features for the dataset are taken into consideration.
	\begin{itemize}
		\item Oblique view. 
		For the UAV platform, it is natural to record images or videos in either an oblique view or a nadir view. Nadir views are common in satellite images as the distance between the camera and the ground is large. Nadir views bring invariance to the representation of objects in the image as only the top of objects can be observed. However, the limited representation also brings confusion in object recognition among different objects. In contrast, an oblique view gives a diverse representation of objects with rich scene context, which can be helpful for object recognition task. When UAV flies closer to the ground, a larger area with more details can be observed, causing large scale variation across an image. In order to observe in an oblique view, the camera angle is set to around 45 degrees to the vertical direction.
		\item High resolution. 
		We adopt 4K resolution video recording mode with safe flying height around 50 meters. The image resolution is either 4096$\times$2160 or 3840$\times$2160. In this setting, it is visually clear enough to differentiate most of the objects. Objects that are horizontally far away could also be detected. In addition, it is even possible to detect humans that are near to the UAV.
		\item Consecutive labeling.  Our dataset is designed for the semantic segmentation task. We prefer to label images in a sequence, where the prediction stability could be evaluated. As it is too expensive to label densely in the temporal space, we label 10 images with 5 seconds interval in each sequence.
		\item Complex and dynamic scenes with diverse objects. 
		Our dataset aims at achieving real-world scene complexity, where there are both static and moving objects. Scenes near streets are chosen for the UAVid dataset as they are complex enough with more dynamic human activities. A variety of objects appear in the scene such as cars, pedestrians, buildings, roads, vegetation, billboards, light poles, traffic lights, and so on. We fly UAVs with an oblique view along the streets or across different street blocks to acquire such scenes.
		\item Data variation. In total, 30 small UAV video sequences are captured in 30 different places to bring variance to the dataset, relieving learning algorithms from over-fitting. Data acquisition is performed in good weather conditions with sufficient illumination. We believe that data acquired in dark environments or other weather conditions like snowing or raining require special processing techniques, which are not the focus of our current dataset. 
	\end{itemize}
	
	DJI phantom3 pro and DJI phantom4 are used for data collection, which are light weighted modern drones. The UAVs fly steadily with a maximum flying speed of 10 m/s, preventing potential blurring effect caused by platform motion. The default cameras mounted on the UAVs are used for video acquisition with only RGB channels.

\subsection{Scene Complexity}
\begin{figure*}[h]%[thpb]
\centering
%%\textcolor{red}{uncomment for image}
\includegraphics[width=1.0\textwidth]{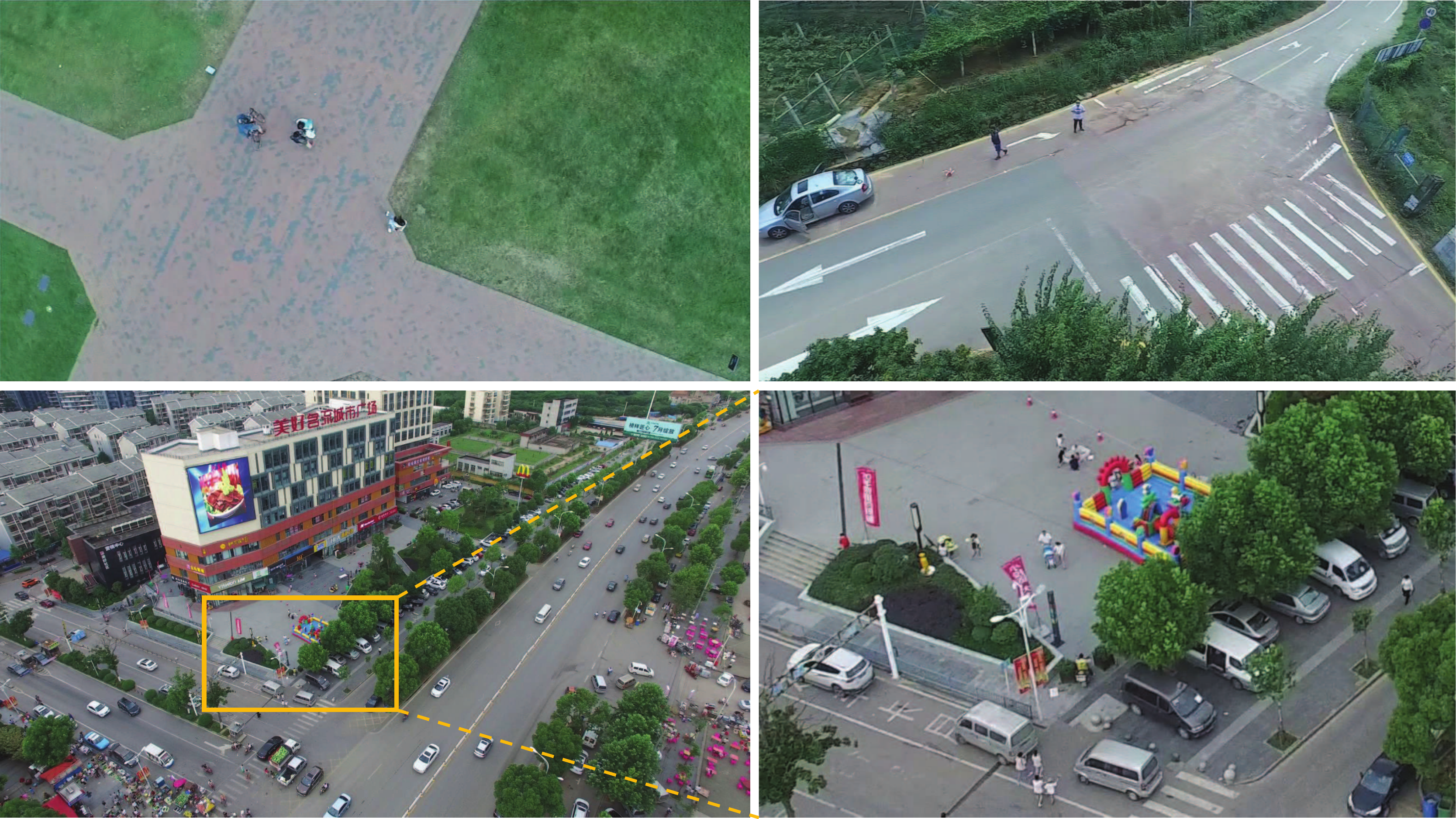}
\caption{Comparison between the Aeroscapes dataset~\cite{aeroscapes} and the UAVid dataset. The first row shows the examples from Aeroscapes dataset. The second row shows the examples from the UAVid dataset, in which the right column shows an image crop at the original scale, where detailed object can be clearly seen. Regarding the number of objects and different object configurations, the UAVid dataset has higher scene complexity.}
\label{fig_ds_cmp}
\end{figure*}

The scene complexity~\cite{Cordts2016Cityscapes} of the new UAVid dataset is higher than the other existing UAV semantic segmentation dataset~\cite{aeroscapes} regarding the number of objects and the different object configurations. We should note that both datasets lack the instance labeling for the quantitative scene complexity calculation~\cite{Cordts2016Cityscapes}. It is still qualitatively evident that our UAVid dataset has much higher scene complexity. We have, on average, $9$ times more car objects and $3$ times more human objects per unit of image area by manually counting in a random subset of images from the two datasets. Examples of the street scenes are shown in Fig.~\ref{fig_ds_cmp}.

\subsection{Dataset Size}
Our UAVid dataset has $300$ images and each of size $4096\times2160$ or $3840\times2160$. To compare the sizes of different datasets for semantic segmentation fairly, we should consider not only the number of images, but also the size of each image. A more fair metric is to compare the number of labeled pixels in total. We select several well-known semantic segmentation datasets for comparisons. The CamVid dataset~\cite{CamVid} has $701$ images of size $960\times720$, which is only one fifth of our dataset in terms of the number of labeled pixels. The giant Cityscapes dataset~\cite{Cordts2016Cityscapes} has $5,000$ images of size $2048\times1024$, which is $4$ times the size of our UAVid dataset. However, many objects in our images are smaller than theirs, providing more object variance in the same number of pixels, which compensate for the object recognition task in a degree. Compared to the ISPRS 2D semantic labeling datasets, the Vaihingen and the Potsdam datasets~\cite{ISPRSbenchmark14} have even fewer images, $33$ and $38$ images respectively, but the size of each image is quite large, e.g., $6000\times6000$ for the Potsdam dataset. Regarding the total number of the labeled pixels, the Vaihingen and the Potsdam datasets are only one tenth and one half the size of our UAVid dataset, respectively. The state-of-the-art DeepGlobe Land Cover Classification dataset~\cite{deepglobe} has $1146$ satellite images for rural areas of size $2448\times2448$, which is about $2.5$ times the size of our dataset. However, the scene complexity of the rural area is much lower than it is in our UAVid dataset. In conclusion, our UAVid dataset has a moderate size, and it is bigger than several well-known semantic segmentation datasets. Section~\ref{sec:exp} has shown that deep learning methods can achieve satisfactory qualitative and quantitative results for experimental purposes, which further proves the usability of our UAVid dataset.

\subsection{Class Definition and Statistical Analysis} ~\label{sec:class_def}
\begin{figure*}[h]%[thpb]
\centering
%\textcolor{red}{uncomment for image}
\noindent\includegraphics[width=1.0\textwidth]{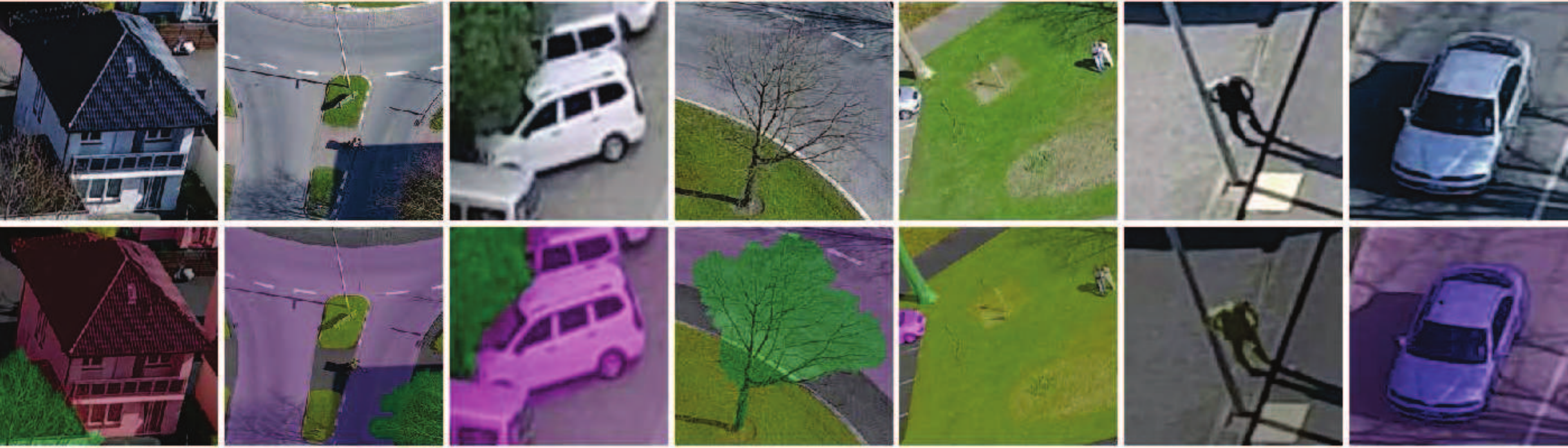}
\caption{\textbf{Example instances from different classes.} The first row shows the cropped instances. The second row shows the corresponding labels. From left to right, the instances are building, road, static car, tree, low vegetation, human and moving car respectively.}
\label{fig_instances}
\end{figure*}

Fully label all types of objects in the street scene in a 4K UAV image is very expensive. As a consequence, only the most common and representative types of objects are labeled for our current dataset. In total, 8 classes are selected for the semantic segmentation, i.e., building, road, tree, low vegetation, static car, moving car, human, and clutter.
Example instances from different classes are shown in Fig.~\ref{fig_instances}.
The definition of each class is described as follows.
\begin{enumerate}%[(1)]
\item building: living houses, garages, skyscrapers, security booths, and buildings under construction. Freestanding walls and fences are not included.
\item road: road or bridge surface that cars can run on legally. Parking lots are not included.
\item tree: tall trees that have canopies and main trunks.
\item low vegetation: grass, bushes and shrubs.
\item static car: cars that are not moving, including static buses, trucks, automobiles, and tractors. Bicycles and motorcycles are not included.
\item moving car: cars that are moving, including moving buses, trucks, automobiles, and tractors. Bicycles and motorcycles are not included.
\item human: pedestrians, bikers, and all other humans occupied by different activities.
\item clutter: all objects not belonging to any of the classes above. 
\end{enumerate}
We deliberately divide the car class into moving car and static car classes. Moving car is such a special class designed for moving object segmentation. Other classes can be inferred from their appearance and context, while the moving car class may need additional temporal information in order to be appropriately separated from static car class. Achieving high accuracy for both static and moving car classes is one possible research goal for our dataset.

\begin{figure*}[h]%[thpb]
\centering
%\textcolor{red}{uncomment for image}
\noindent\includegraphics[width=1.0\textwidth]{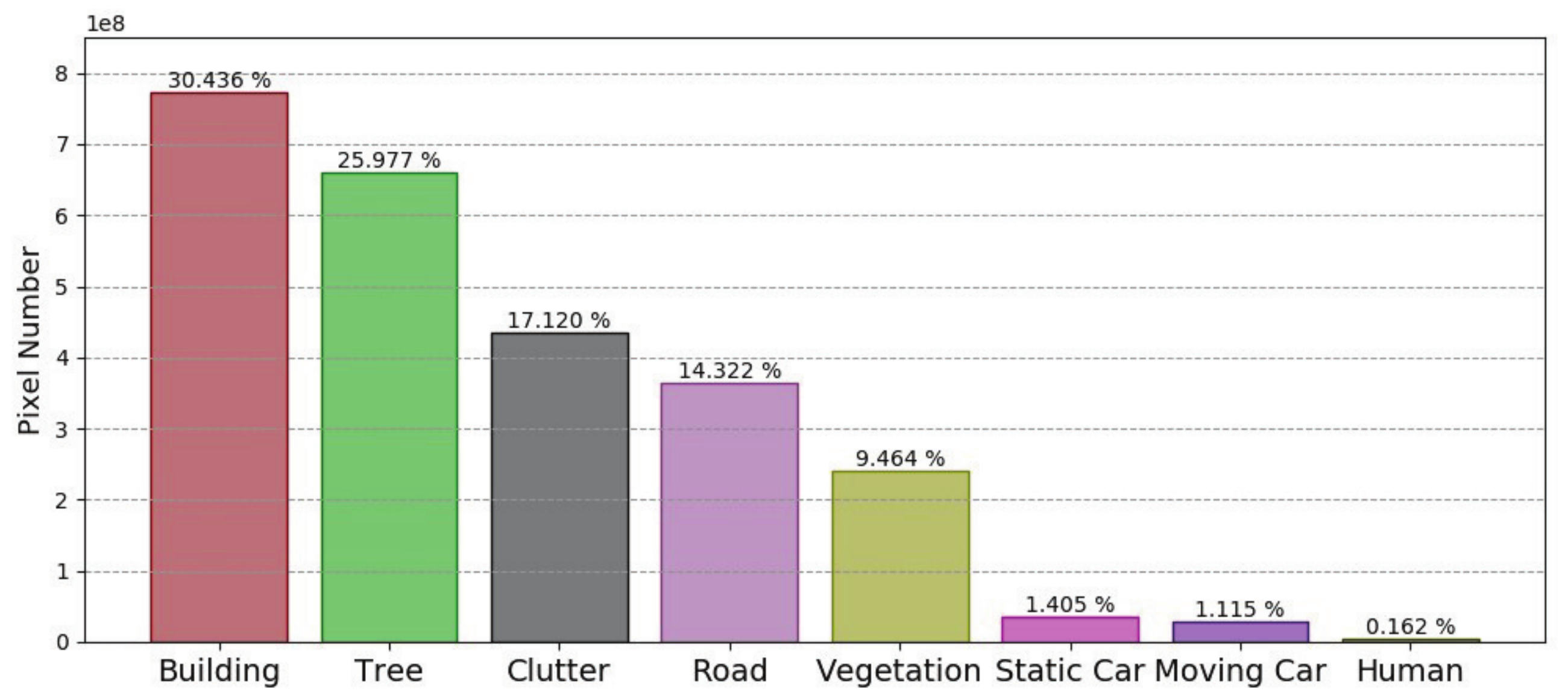}
\caption{\textbf{Pixel number histogram.}}
\label{fig_statistics}
\end{figure*}

The number of pixels in each of the $8$ classes from all $30$ sequences is reported in Fig.~\ref{fig_statistics}. It clearly shows the unbalanced pixel number distribution of different classes. Most of the pixels are from classes like building, tree, clutter, road, and low vegetation. Fewer pixels are from moving car and static car classes, which are both fewer than 2\% of the total pixels. For human class, it is almost zero, fewer than 0.2\% of the total pixels. Smaller pixel number is not necessarily resulted by fewer instances, but the size of each instance. A single building can take more than 10k pixels, while a human instance in the image may only take fewer than 100 pixels. Normally, classes with too small pixel numbers are ignored in both training and evaluation for semantic segmentation task~\cite{Cordts2016Cityscapes}. However, we believe humans and cars are important classes that should be kept in street scenes rather than being ignored.

\subsection{Annotation Method} ~\label{sec:anno}
\begin{figure*}[h]%[thpb]
\centering
%%\textcolor{red}{uncomment for image}
\includegraphics[width=1.0\textwidth]{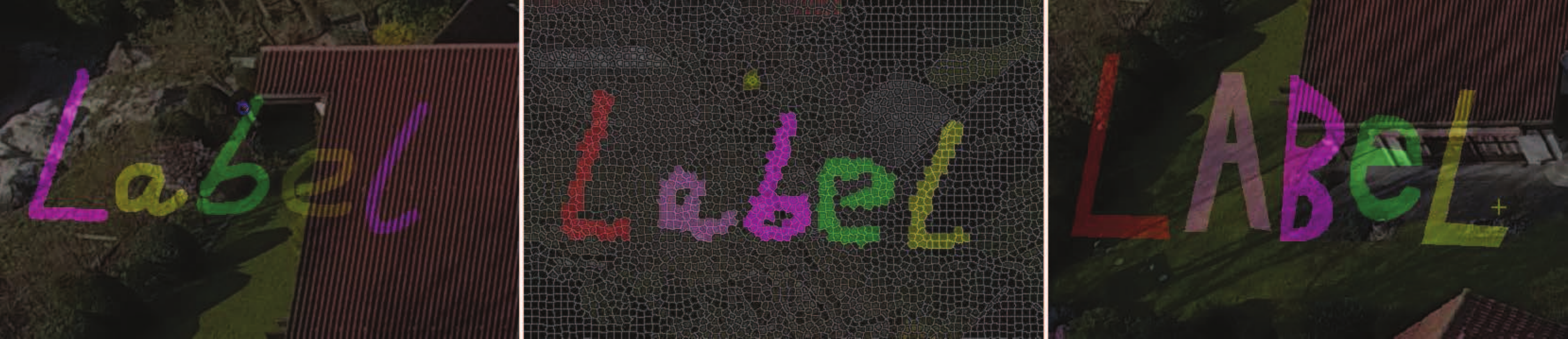}
\caption{\textbf{Annotation methods}. Left shows pixel level annotation, middle shows super-pixel level annotation, and right shows polygon level annotation.}
\label{fig_annotation}
\end{figure*}

We have provided densely labeled fine annotations for high-resolution UAV images. All the labels are acquired with our own labeler tool. It takes approximately $2$ hours to label all pixels in one image.
Pixel level, super-pixel level, and polygon level annotation methods are provided for annotators, as illustrated in Fig.\ref{fig_annotation}. For super-pixel level annotation, our method employs a similar strategy as the COCO-Stuff~\cite{coco_stuff} dataset. We first apply SLIC method~\cite{slic} to partition the image into super-pixels, each of which is a group of pixels that are spatially connected and share similar characteristics, such as color and texture. The pixels within the same super-pixel are labeled with the same class. Super-pixel level annotation can be useful for the objects with sawtooth boundaries like trees. We offer super-pixel segmentation of $4$ different scales for annotators to best adjust to objects of different scales. Polygon annotation is more useful to annotate objects with straight boundaries like buildings, while pixel level annotation serves as a basic annotation method. Our tool also provides video play functionality around certain frames to help to inspect whether certain objects are moving or not. As there might be overlapping objects, we label the overlapping pixels to be the class that is closer to the camera.

\subsection{Dataset Splits} ~\label{sec:ds_split}
The whole 30 densely labeled video sequences are divided into training, validation, and test splits. We do not split the data completely randomly, but in a way that makes each split to be representative enough for the variability of different scenes. All three splits should contain all classes. Our data is split at the sequence level, and each sequence comes from a different scene place. Following this scheme, we get 15 training sequences (150 labeled images) and 5 validation sequences (50 labeled images) for training and validation splits, respectively, whose annotations will be made publicly available. The test split consists of the left 10 sequences (100 labeled images), whose labels are withheld for benchmarking purposes. The size ratios among training, validation and test splits are 3:1:2.

\section{Semantic Labeling Task} ~\label{sec:task}
In this section, the semantic labeling task for our dataset is introduced. The task details and the evaluation metric for the UAVid dataset are introduced first in section~\ref{sec:T&M}. The following sections (from ~\ref{sec:bl_dnn} to~\ref{sec:bl_reg}) introduce the baseline methods for the task. The baseline methods are presented in company with the task to offer performance references and to test the usability of the dataset for the task. Section~\ref{sec:bl_dnn} and section~\ref{sec:bl_msd} introduce the deep neural networks in the baseline methods. Section~\ref{sec:bl_ft} and section~\ref{sec:bl_reg} introduce the pre-training and the spatial-temporal regularization respectively, which boost the performance for all baseline methods.

\subsection{Task and Metric} \label{sec:T&M}
The task defined on the UAVid dataset is to predict pixel level semantic labeling for the UAV images. The image-label pairs are provided for each sequence together with the unlabeled images. Currently, the UAVid dataset only supports image level semantic labeling without instance level consideration.
The semantic labeling performance is assessed based on the standard mean IoU metric~\cite{PascalVOC}. The goal for this task is to achieve as a high mean IoU score as possible. For the UAVid dataset, the clutter class has a relatively large pixel number ratio and consists of meaningful objects, which is taken as one class for both training and evaluation rather than being ignored.

\subsection{Deep neural networks for baselines} \label{sec:bl_dnn}
In order to offer performance references and to test the usability of our UAVid dataset for the semantic labeling task, we have tested several deep learning models for the single image prediction. Although static cars and moving cars cannot be differentiated by their appearance from only one image, it is still possible to infer based on their context. The moving cars are more likely to appear in the center of the road, while the static cars are more likely to be at the parking lots or to the side of the roads.
As the UAVid dataset consists of very complex street scenes, it requires powerful algorithms like deep neural networks for the semantic labeling task. 
We start with 3 widely used deep fully convolutional neural networks, they are FCN-8s~\cite{FCN8s}, Dilation net~\cite{dilationNet} and U-Net~\cite{UNet}.

FCN-8s~\cite{FCN8s} has often been a good baseline candidate for semantic segmentation. It is a giant model with strong and effective feature extraction ability, but yet simple in structure. It takes a series of simple 3x3 convolutional layers to form the main parts for high-level semantic information extraction. This simplicity in structure also makes FCN-8s popular and widely used for semantic segmentation.

Dilation net~\cite{dilationNet} has a similar front end structure with FCN-8s, but it removes the last two pooling layers in VGG16. Instead, convolutions in all following layers from the conv5 block are dilated by a factor of 2 due to the ablated pooling layers. Dilation net also applies a multi-scale context aggregation module in the end, which expands the receptive field to boost the performance for prediction. The module is achieved by using a series of dilated convolutional layers, whose dilation rate gradually expands as the layer goes deeper.

U-Net~\cite{UNet} is a typical symmetric encoder-decoder network originally designed for segmentation on medical images. The encoder extracts features, which are gradually decoded through the decoder. The features from each convolutional block in the encoder are concatenated to the corresponding convolutional block in the decoder to acquire features of higher and higher resolution for prediction gradually. U-Net is also simple in structure but good at preserving object boundaries.

\begin{figure*}[ht]
%\textcolor{red}{uncomment for image}
\centering
\noindent
\includegraphics[width=1.0\textwidth]{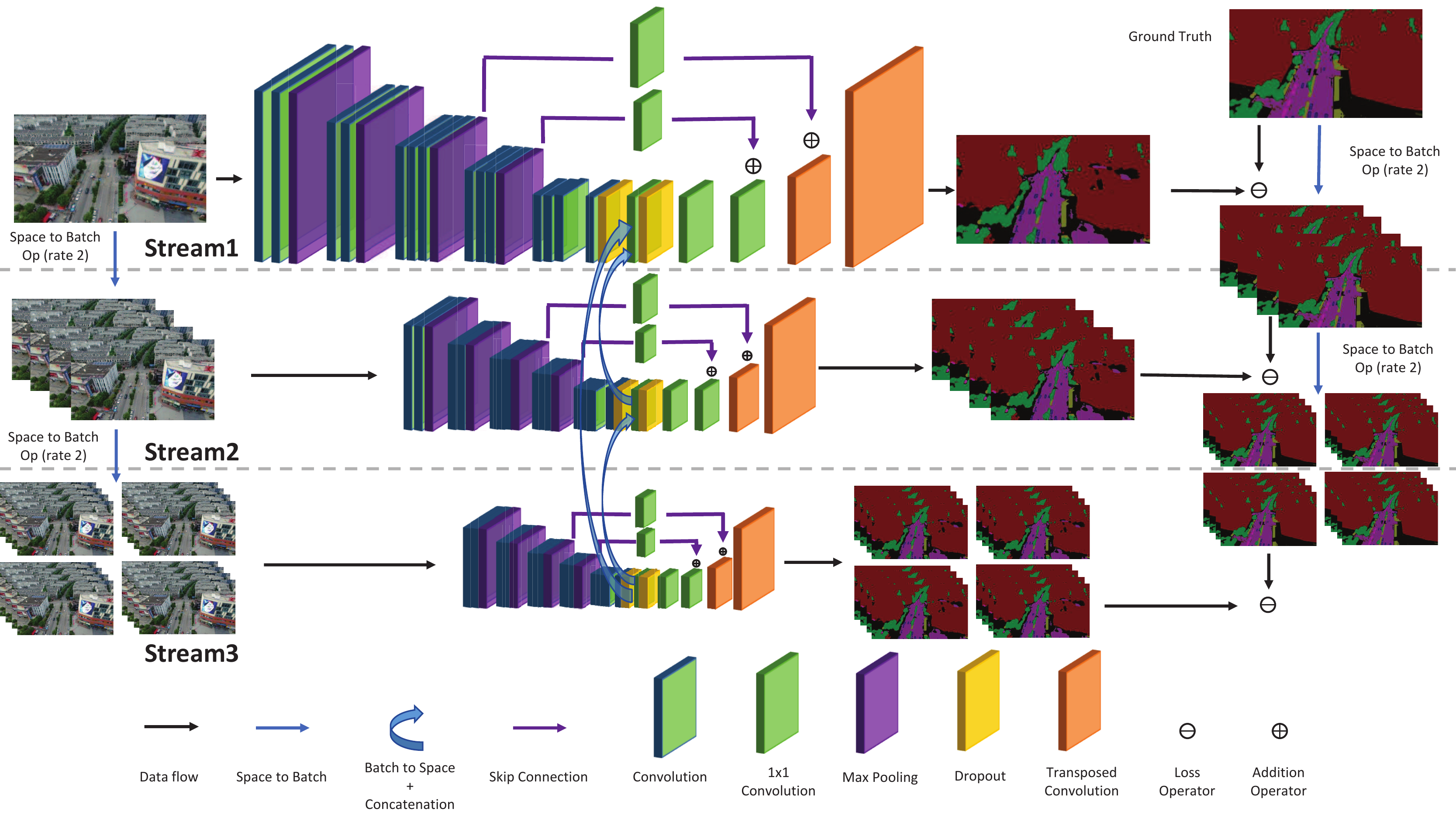}
\caption{\textbf{Structure of the proposed Multi-Scale-Dilation network.} Three scales of images are achieved by Space to Batch operation with rate 2. Standard convolutions in stream2 and stream3 are equivalent to dilated convolutions in stream1. The main structure for each stream is FCN-8s~\cite{FCN8s}, which could be replaced by any other networks. Features are aggregated at conv7 layer for better prediction on finer scales.}
\label{fig_fcn_ms}
\end{figure*}

\subsection{Multi-Scale-Dilation Net} \label{sec:bl_msd}
\begin{figure*}[ht]
%\textcolor{red}{uncomment for image}
\centering
\noindent
\includegraphics[width=1.0\textwidth]{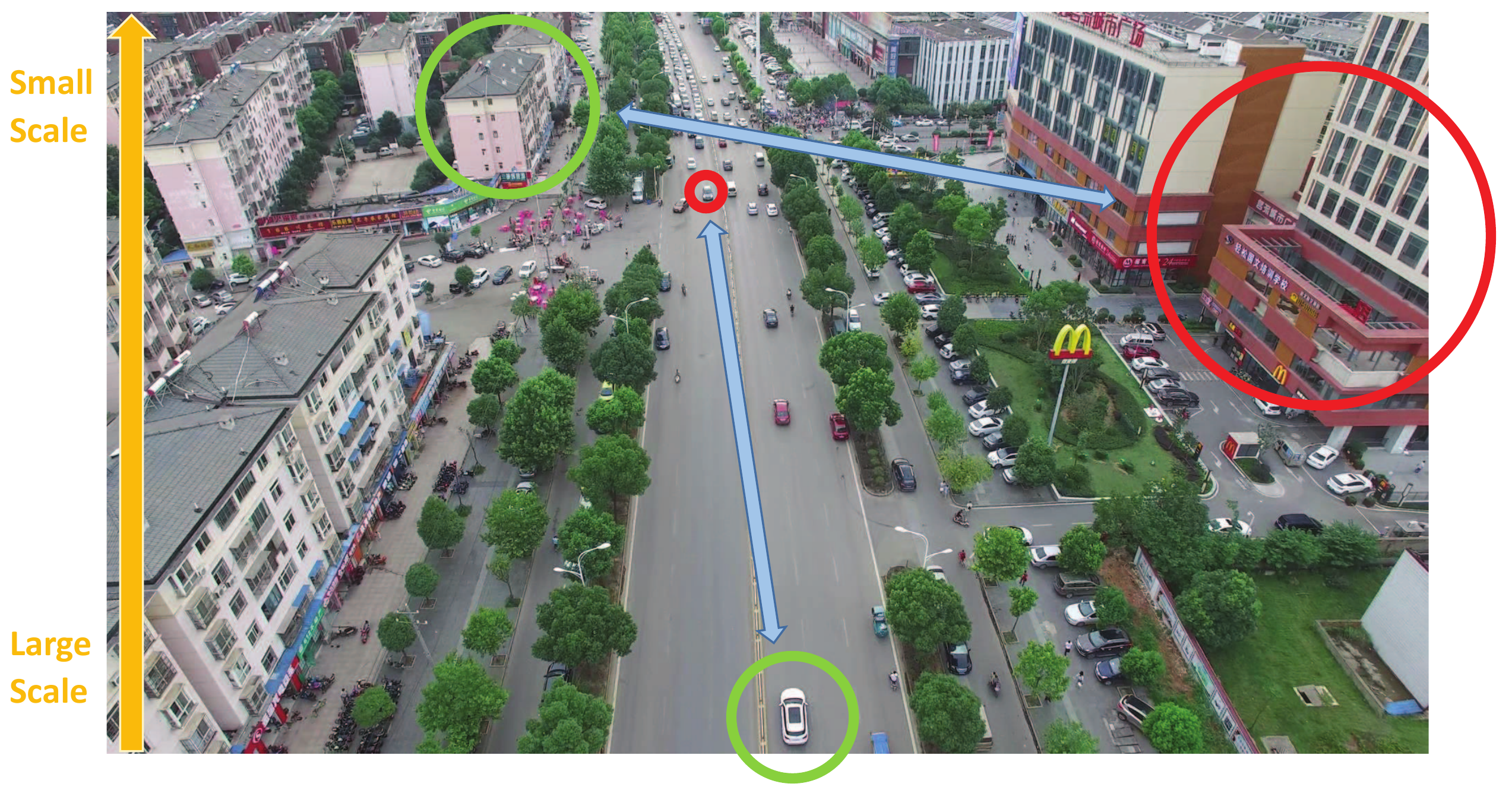}
\caption{Illustration of the scale problem in an UAV image. The scales of the objects vary greatly from the bottom to the top of the image. The green circles mark the objects in proper scales while the red circles mark the objects in either too large or too small scales.}
\label{fig_scale_problem}
\end{figure*}

For a high-resolution image captured by a UAV in a slanted view, the sizes of objects in different distances can vary dramatically. Figure~\ref{fig_scale_problem} illustrates such a scale problem in the UAVid dataset.
The large scale variation in a UAV image can affect the accuracy of prediction. In a network, each output pixel in the final prediction layer has a fixed receptive field, which is formed by pixels in the original image that can affect the final prediction of that output pixel. When the objects are too small, the neural network may learn the noise from the background. When the objects are too big, the model may not acquire enough information to infer the label correctly. The above is a long-standing notorious problem in computer vision. To reduce such a large scale variation effect, we propose a novel multi-scale-dilation net (MS-Dilation net) as an additional baseline.

One way to expand the receptive field of a network is to use dilated convolution~\cite{dilationNet}. Dilated convolution can be implemented in different ways, one of which is to leverage on space to batch operation (S2B) and batch to space operation (B2S), which is provided in Tensorflow API. Space to batch operation outputs a copy of the input tensor where values from the height and width dimensions are moved to the batch dimension. Batch to space operation does the inverse. A standard 2D convolution on the image after S2B is the same as a dilated convolution on the original image. 
A single dilated convolution can be performed as $S2B->convolution->B2S$. 
This implementation for dilated convolution is efficient when there is a cascade of dilated convolutions, where intermediate S2B and B2S cancel out. For instance, 2 consecutive dilated convolution with the same dilation rate can be performed as $S2B->convolution->convolution->B2S$.

Space to batch operation can also be taken as a kind of nearest neighbor down-sampling operation, where the input is the original image while the outputs are down-sampled images with slightly different spatial shifts. The nearest neighbor down-sampling operation is nearly equivalent to space to batch operation, where the only difference is the number of output batches. With the above illustration, it is easy to draw the connection between the dilated convolution and the standard convolution on down-sampled images. 

By utilizing space to batch operation and batch to space operation, semantic segmentation can be done on different scales. In total, three streams are created for three scales, as shown in Fig.~\ref{fig_fcn_ms}. For each stream, a modified FCN-8s is used as the main structure, where the depth for each convolutional block is reduced due to the memory limitation. Here, filter depth is sacrificed for more scales. In order to reduce detail loss in feature extraction, the pooling layer in the fifth convolutional block is removed to keep a smaller receptive field. Instead, features with larger receptive fields from other streams are concatenated to higher resolution features through skip connection in conv7 layers. Note that these skip connections need batch to space operation to retain spatial and batch number alignment. In this way, each stream handles feature extraction in its own scale, and features from larger scales are aggregated to boost prediction for higher resolution streams.

Multiple scales may also be achieved by down-sampling images directly~\cite{pyramid_images}. However, there are 3 advantages to our multi-scale processing. First, every pixel is assigned to one batch in space to batch operation, and all the labeled pixels shall be used for each scale with no waste. Second, there is strict alignment between image-label pairs in each scale as there is no mixture of image pixels nor a mixture of label pixels. Finally, the concatenated features in the conv7 layer are also strictly aligned.

For each scale, corresponding ground truth labels can also be generated through space to batch operation in the same way as the generation for input images in different streams. With ground truth labels for each scale, deeply supervised training can be done. The losses in three scales are all cross entropy loss. The loss in stream1 is the target loss, while the losses in stream2 and stream3 are auxiliary. The final loss to be optimized is the weighted mean of the three losses, shown in the equation below. $m1,m2,m3$ are the numbers of pixels of an image in each stream. $n$ is the batch index, and $t$ is the pixel index. $p$ is the target probability distribution of a pixel, while $q$ is the predicted probability distribution.

$$CE_1 = \frac{1}{m_1}\sum_{t=1}^{m_1}-p_t\log(q_t)\eqno{(1)}$$
$$CE_2 = \frac{1}{4m_2}\sum_{n=1}^{4}\sum_{t=1}^{m_2}-p_t^n\log(q_t^n)\eqno{(2)}$$
$$CE_3 = \frac{1}{16m_3}\sum_{n=1}^{16}\sum_{t=1}^{m_3}-p_t^n\log(q_t^n)\eqno{(3)}$$
$$Loss = \frac{w_1 \times CE_1 + w_2 \times CE_2 + w_3 \times CE_3}{w_1 + w_2 + w_3}\eqno{(4)}$$

It is also interesting to note that every layer becomes a dilated version for stream2 and stream3, especially for the pooling layer and the transposed convolutional layer, which turn into a dilated pooling layer and a dilated transposed convolutional layer respectively. Compared to layers in stream1, layers in stream2 are dilated by rate of 2, and layers in stream3 are dilated by rate of 4. Theses 3 streams together form the MS-Dilation net.

\subsection{Fine-tune Pre-trained Networks} \label{sec:bl_ft}
Due to the limited size of our UAVid dataset, training from scratch may not be enough for the networks to learn diverse features for better label prediction. Pre-training a network has been proved to be very useful for various benchmarks~\cite{pretrain_isprs,pretrain_davis,pretrain_cityscape,pretrain_ade20k}, which boosts the performance by utilizing more data from other datasets. To reduce the effect of limited training samples, we also explore how much pre-training a network can boost the score for the UAVid semantic labeling task. We pre-train all the networks with the cityscapes dataset~\cite{Cordts2016Cityscapes}, which comprises many more images for training.

\begin{figure*}[ht]
%\textcolor{red}{uncomment for image}
\centering
\noindent
\includegraphics[width=1.0\textwidth]{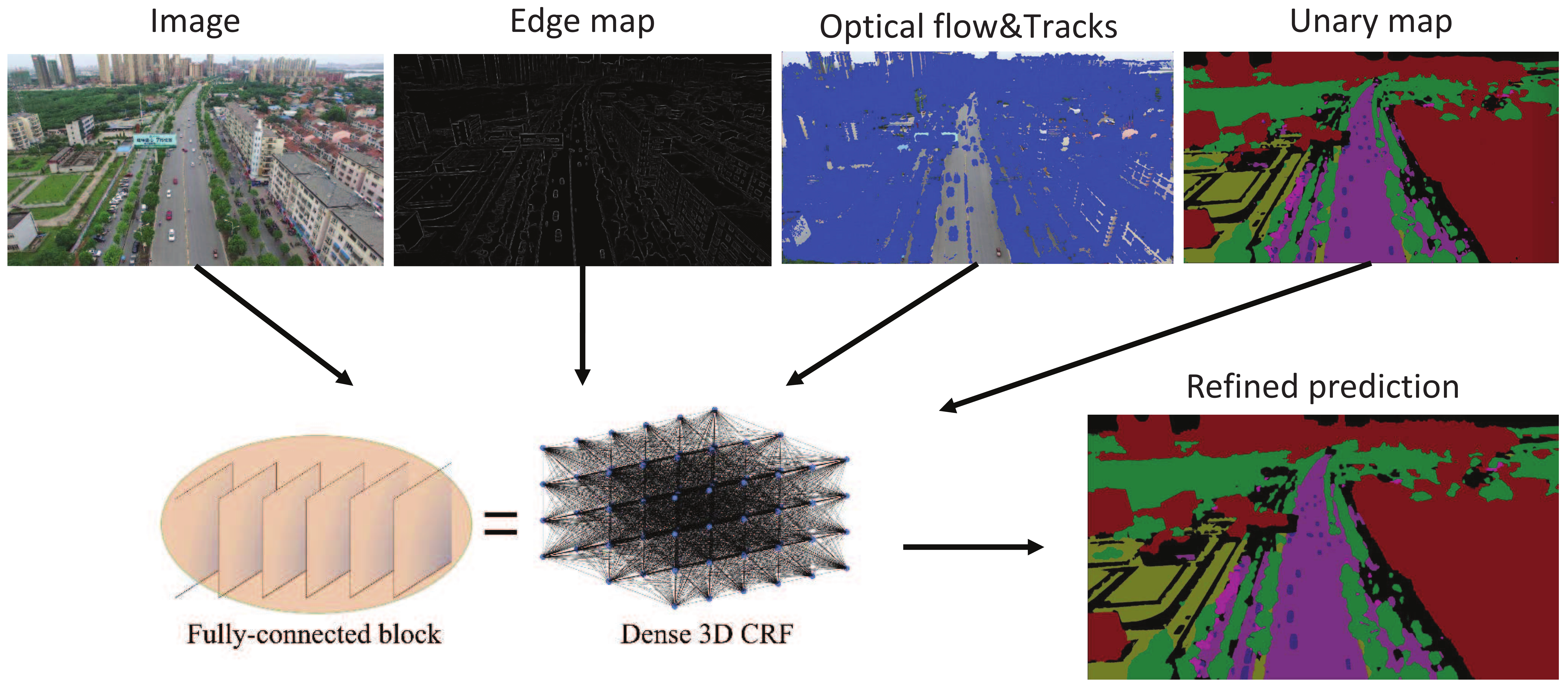}
\caption{The data inputs for the FSO~\cite{FSO} post-processing method. The image, the edge map, the unary map, the optical flow and tracks are required for the method. The edge map shows the probability of each pixel to be an edge. The blue points in the image of optical flow\&Tracks mark the points being tracked. The unary map is the class probabilities for each pixel predicted by the deep neural networks.}
\label{fig_fso_ingredient}
\end{figure*}

\subsection{Spatial-temporal Regularization for Semantic Segmentation} \label{sec:bl_reg}
For semantic labeling task, we further explore how a spatial-temporal regularization can improve the prediction. Taking advantage of temporal information is valuable for label prediction for sequence data. Normally, deep neural networks trained on individual images cannot provide completely consistent predictions spanning several frames. However, different frames provide observations from different viewing positions, through which multiple clues can be collected for object prediction. To utilize temporal information in the UAVid dataset, we adopt feature space optimization (FSO)~\cite{FSO} method for sequence data prediction. It smooths the final label prediction for the whole sequence by applying 3D CRF covering both spatial and temporal domains. The method takes advantage of optical flow and tracks to link the pixels in the temporal domain. The whole post-processing requires multiple data inputs, including the image, the unary map from the deep neural networks, the edge map, the optical flow, and the tracks as shown in Fig.~\ref{fig_fso_ingredient}.

\section{Experiments} ~\label{sec:exp}
Our experiments are divided into 3 parts. Firstly, we compare semantic segmentation results by training deep neural networks from scratch. These results serve as the basic baselines. Secondly, we analyze how pre-trained models can be useful for the UAVid semantic labeling task. We fine-tune deep neural networks with UAVid dataset after they are pre-trained on the cityscapes dataset~\cite{Cordts2016Cityscapes}. Finally, we explore the influence of spatial-temporal regulation by using the FSO method for semantic video segmentation.

\begin{figure*}%[thpb]
\centering
%\textcolor{red}{uncomment for image}
\includegraphics[width=1.0\textwidth]{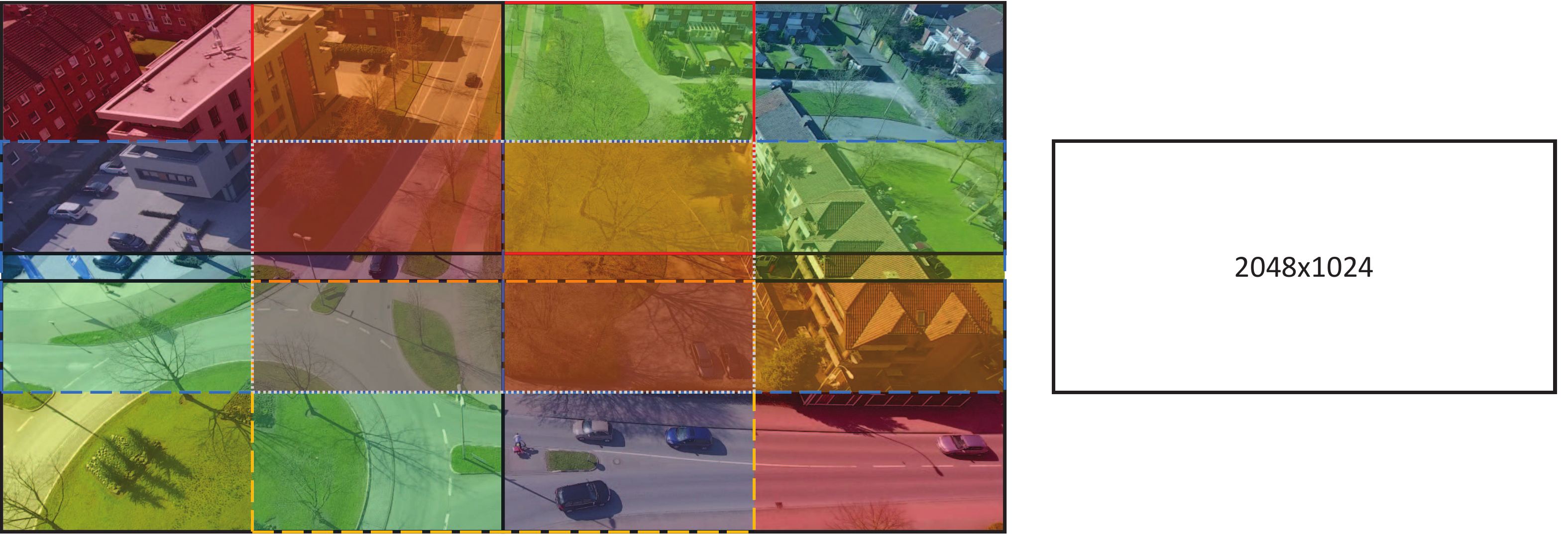}
\caption{Image cropping illustration. The 4K image is cropped to 9 evenly distributed smaller overlapped images before processing.}
\label{fig:img_clip}
\end{figure*}

The size of our UAV images is very large, which requires too much GPU memory for intermediate feature storage in deep neural networks. As a result, we crop each UAV image into 9 evenly distributed smaller overlapped images that cover the whole image for training. Each cropped image is of size 2048$\times$1024. We keep such a moderate image size in order to reduce the ratio between the zero padding area and the valid image area. Bigger image size also resembles a larger batch size when each pixel is taken as a training sample. During testing, the average prediction scores are used for the overlapped area. Fig.~\ref{fig:img_clip} illustrates the way of cropping.

\begin{table*}
\centering
\caption{\textbf{IoU scores for different models.} IoU scores are reported in percentage and best results are shown in bold. \textbf{PRT} stands for pre-train and \textbf{FSO} stands for feature space optimization~\cite{FSO}.}
\label{tb_IoU}
\resizebox{2.0\columnwidth}{!}{%
	\begin{tabular}{lccccccccc}
		\hline
		%		\bf Model &\bf {BD}&\bf {TR}&\bf {CLT}&\bf {RD}&\bf {VG}&\bf {SC}&\bf {MC}&\bf {HM}&\bf {Mean IoU}\\
		\bf Model &\bf {Building}&\bf {Tree}&\bf {Clutter}&\bf {Road}&\bf {Low Vegetation}&\bf {Static Car}&\bf {Moving Car}&\bf {Human}&\bf {mean IoU}\\
		\hline
		FCN-8s					     &   64.3&   63.8&   33.5&   57.6&   28.1&    8.4&   29.1&   0.0&   35.6\\
		%\hline
		Dilation Net			     &   72.8&   66.9&   38.5&   62.4&   34.4&    1.2&   36.8&   0.0&   39.1\\
		%\hline
		U-Net					     &   70.7&   67.2&   36.1&   61.9&   32.8&   11.2&\bf47.5&   0.0&   40.9\\
		%\hline
		MS-Dilation (ours)	     &\bf74.3&\bf68.1&\bf40.3&\bf63.5&\bf35.5&\bf11.9&   42.6&   0.0&\bf42.0\\
		\hline
		FCN-8s+PRT				     &   77.4&   72.7&   44.0&   63.8&   45.0&   19.1&   49.5&   0.6&   46.5\\
		%\hline
		Dilation Net+PRT		     &\bf79.8&   73.6&   44.5&   64.4&   44.6&   24.1&   53.6&   0.0&   48.1\\
		%\hline
		U-Net+PRT				     &   77.5&   73.3&   44.8&   64.2&   42.3&\bf25.8&\bf57.8&   0.0&   48.2\\
		%\hline
		MS-Dilation (ours)+PRT    &   79.7&\bf74.6&\bf44.9&\bf65.9&\bf46.1&   21.8&   57.2&   \bf8.0&\bf49.8\\
		\hline
		FCN-8s+PRT+FSO			     &   78.6&   73.3&   45.3&   64.7&   46.0&   19.7&   49.8&   0.1&   47.2\\
		%\hline
		Dilation Net+PRT+FSO	     &   80.7&   74.0&   45.4&   65.1&   45.5&   24.5&   53.6&   0.0&   48.6\\
		%\hline
		U-Net+PRT+FSO			     &   79.0&   73.8&\bf46.4&   65.3&   43.5&\bf26.8&   56.6&   0.0&   48.9\\
		%\hline
		MS-Dilation (ours)+PRT+FSO&\bf80.9&\bf75.5&   46.3&\bf66.7&\bf47.9&   22.3&\bf56.9&\bf4.2&\bf50.1\\
		\hline
	\end{tabular}
}
\end{table*}

\subsection{Train from Scratch} \label{train_from_scratch}
\begin{figure*}%[thpb]
\centering
%\textcolor{red}{uncomment for image}
\includegraphics[width=1.0\textwidth]{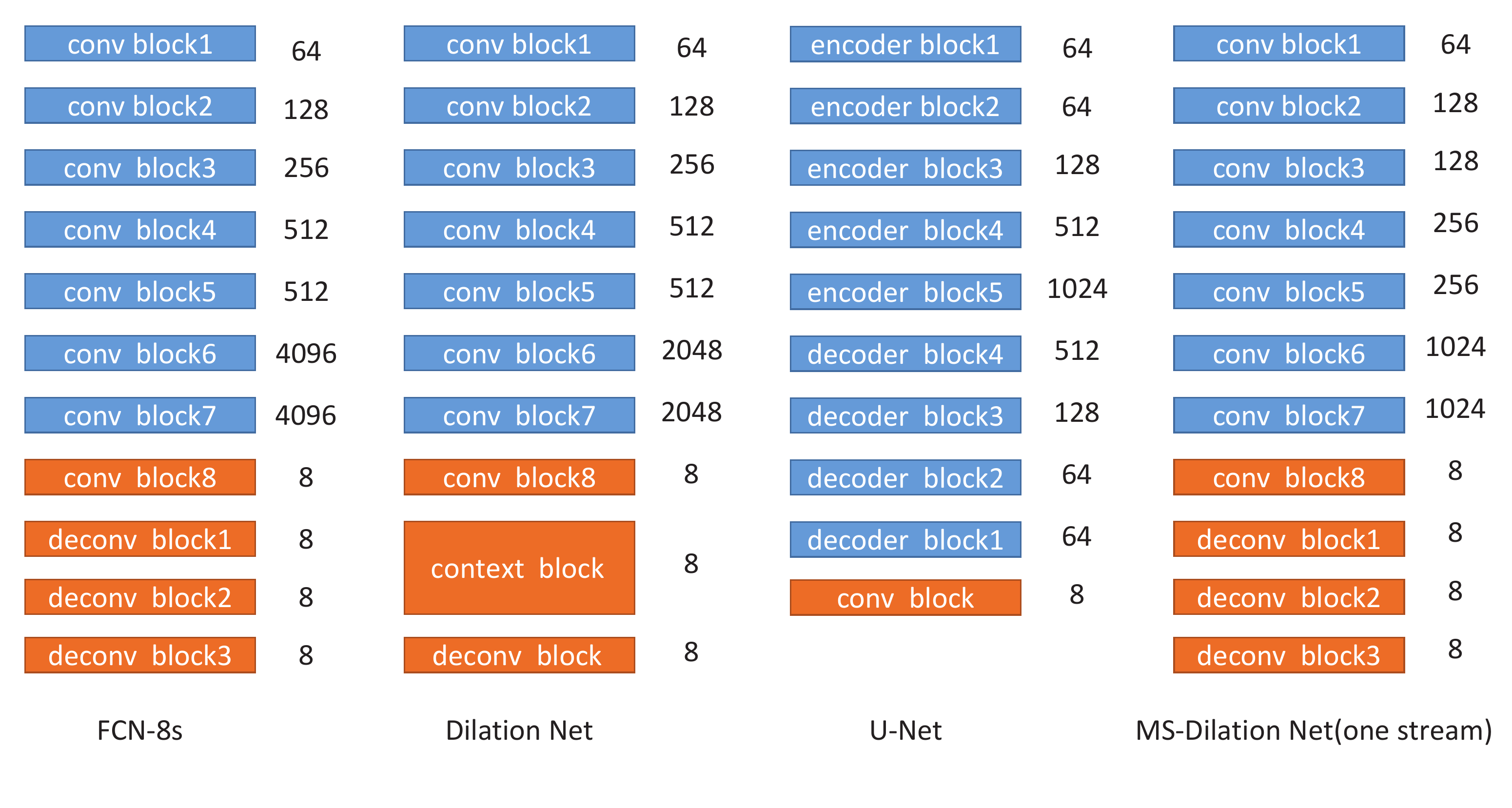}
\caption{The configuration of different models. The blue blocks are the feature extraction part, while the orange blocks are the context aggregation and the prediction part for the corresponding 8 classes in UAVid dataset.}
\label{fig:model_details}
\end{figure*}

To have a fair comparison among different networks, we re-implement all the networks with Tensorflow~\cite{tensorflow}, and all networks are trained with an Nvidia Titan X GPU. In order to accommodate the networks into 12G GPU memory, depth of some layers in the Dilation net, U-Net, and MS-Dilation net are reduced to fit into the memory maximally. The model configuration details of different networks are shown in Fig.~\ref{fig:model_details}.

The neural networks share similar hyper-parameters for training from scratch. All models are trained with the Adam optimizer for 27K iterations (20 epochs). The base learning rate is set to $10^{-4}$ exponentially decaying to $10^{-7}$. Weight decay for all weights in convolutional kernels is set to $10^{-5}$. Training is done with one image per batch. 
For data augmentation in training, we apply left-to-right flip randomly. We also apply a series of color augmentation, including random hue operation, random contrast operation, random brightness operation, random saturation operation.

Auxiliary losses are used for our MS-Dilation net. The loss weights for three streams are set to 1.8, 0.8, and 0.4 empirically. The loss weights for stream2 and stream3 are set smaller than stream1 as the main goal is to minimize the loss in stream1. For the Dilation net, the basic context aggregation module is used and initialized as it is in~\cite{dilationNet}. All networks are trained end-to-end, and their mean IoU scores are reported in percentage, as shown in Tab.~\ref{tb_IoU}.

For all the four networks, they are better at discriminating building, road, and tree classes, achieving IoU scores higher than 50\%. The scores for car, vegetation, and clutter classes are relatively lower. All four networks completely fail to discriminate human class. Normally, classes with larger pixel number have relatively higher IoU scores. However, the IoU score for the moving car class is much higher than the static car class, even though the two classes have similar pixel numbers. The reason may be that static cars may appear in various contexts like parking lots, garages, sidewalks, or partially blocked under the trees, while moving cars often run in the middle of roads with a very clear view.

The Dilation net and the U-Net perform similarly, and they both outperform the FCN-8s. The FCN-8s extracts features on a single scale, while the Dilation net and U-Net benefit from features in better scales from the context blocks and multiple decoders in multiple scales, respectively. Our Multi-Scale-Dilation net differs as it extracts features in multiple scales from very early and shallow layers, and it achieves the best mean IoU score and the best IoU scores for most of the classes among the four networks. It shows the effectiveness of multi-scale feature extraction.

\subsection{Fine-tune Pre-trained Models}
For fine-tuning, all the networks are pre-trained with cityscapes dataset~\cite{Cordts2016Cityscapes}. Finely annotated data from both training and validation splits are used, which is of 3,450 densely labeled images in total. Hyper-parameters and data augmentation are set the same as they are in section~\ref{train_from_scratch}, except that the iteration is set to 52K. Next, all the networks are fine-tuned with data from the UAVid dataset. As there is still large heterogeneity between these two datasets, all layers are trained for all networks. We only initialize feature extraction parts of the networks with pre-trained models, while the prediction parts are initialized the same as training from scratch. The learning rate is set to $10^{-5}$ exponentially decaying to $10^{-7}$ for FCN-8s, and $10^{-4}$ exponentially decaying to $10^{-7}$ for other 3 networks as they are more easily stuck at a local minimum with initial learning rate to be $10^{-5}$ during training. The rest of the hyper-parameters are set the same as training from scratch. The performance is also shown in Tab.~\ref{tb_IoU}.

To find out whether auxiliary losses are important, we have fine-tuned MS-Dilation net with 3 different training plans. For the first plan, we fine-tune the MS-Dilation net without auxiliary losses for 30 epochs by setting loss weights to 0 in stream2 and stream3. For the second plan, we fine-tune the MS-Dilation net with auxiliary losses for 30 epochs. For the final plan, we fine-tune the MS-Dilation net with auxiliary losses for 20 epochs and without auxiliary losses for another 10 epochs. The IoU scores for three plans are shown in Tab.~\ref{tb_IoU_ms}. As it is shown, the best mean IoU score is achieved by the third plan. The better result for MS-Dilation net+PRT in Tab.~\ref{tb_IoU} is achieved by fine-tuning 20 epochs without auxiliary losses after fine-tuning 20 epochs with auxiliary losses.

Clearly, auxiliary losses are very important for the MS-Dilation net. However, neither purely fine-tuning the MS-Dilation net with auxiliary losses nor without achieves the best score. It is the combination of these two fine-tuning processes that brings the best score. 
Auxiliary losses are important as they can guide the multi-scale feature learning process, but the network needs to be further fine-tuned without auxiliary losses to get the best multi-scale filters for prediction.

By fine-tuning the pre-trained models, the performance boost is huge for all networks across all classes except human class. The networks still struggle to differentiate human class. Nevertheless, the improvement is evident for the MS-Dilation net with 8\% improvement. Decoupling the filters with different scales can be beneficial when objects appear in large scale variation.

\begin{figure*}%[thpb]
\centering
%\textcolor{red}{uncomment for image}
\includegraphics[width=1.0\textwidth]{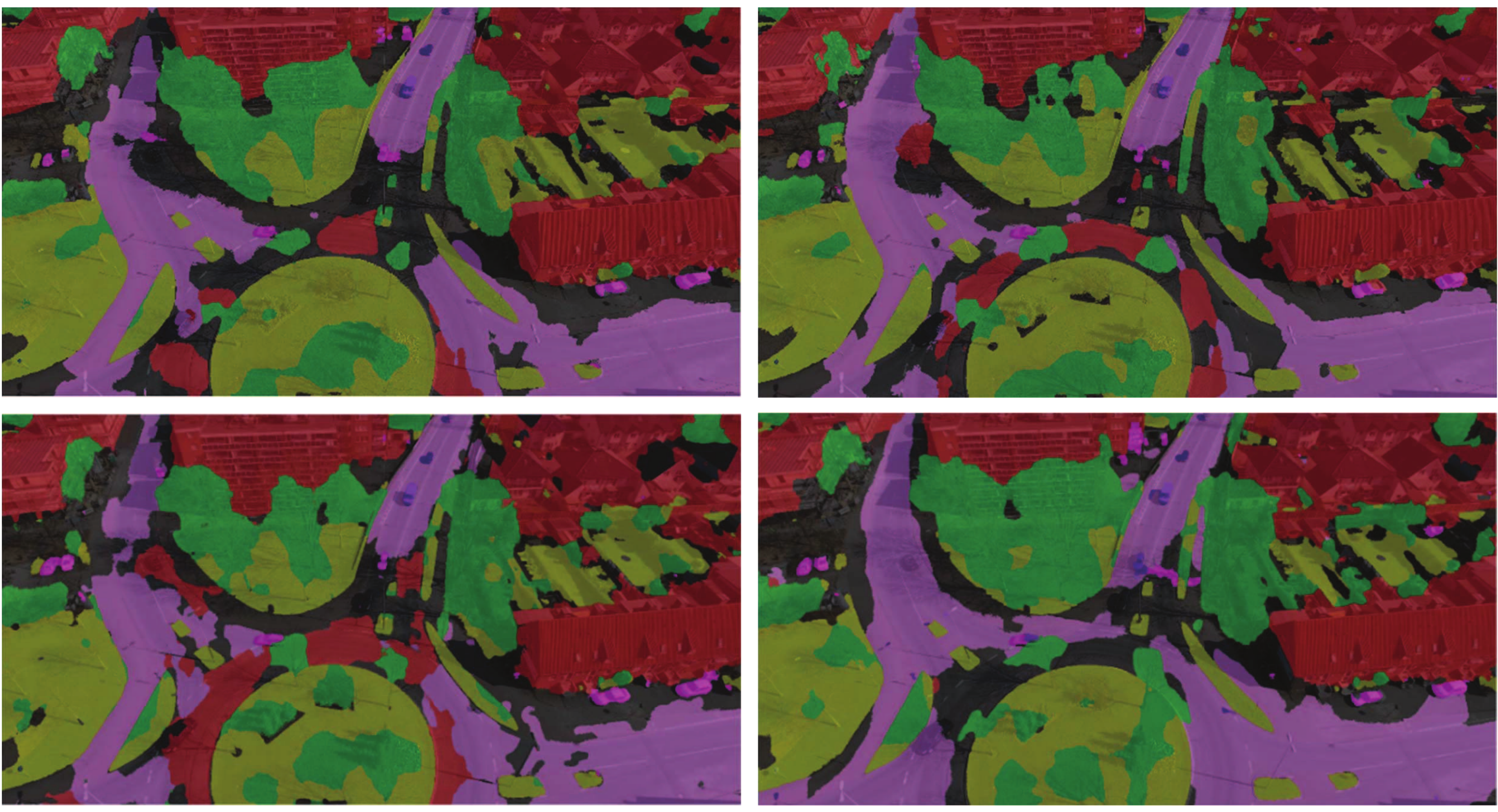}
\includegraphics[width=0.80\textwidth]{./legends}
\caption{Prediction example of FCN8s (top left), Dilation Net (top right), U-Net (bottom left) and MS-Dilation Net (bottom right).}
\label{fig_ms_eg}
\end{figure*}

In order to see the effect of multiple-scale processing, the qualitative performance comparisons among FCN-8s, Dilation net, U-Net, and MS-Dilation Net are presented in Fig.~\ref{fig_ms_eg}. By utilizing features in multiple scales, the MS-Dilation Net gives relatively better prediction for the roundabout. Locally, the road may be wrongly classified to be building due to the simple texture. However, by aggregating information from multiple scales in MS-Dilation Net, the relatively better label can be predicted.

\begin{table*}
\centering
\caption{\textbf{IoU scores for different training strategies.} IoU scores are reported in percentage and best results are shown in bold. \textbf{w} stands for with and \textbf{w/o} stands for without.}
\label{tb_IoU_ms}
\resizebox{2.0\columnwidth}{!}{%
	\begin{tabular}{lccccccccc}
		\hline
		%		\bf Method &\bf {BD}&\bf {TR}&\bf {CLT}&\bf {RD}&\bf {VG}&\bf {SC}&\bf {MC}&\bf {HM}&\bf {Mean IoU}\\
		\bf Model &\bf {Building}&\bf {Tree}&\bf {Clutter}&\bf {Road}&\bf {Low Vegetation}&\bf {Static Car}&\bf {Moving Car}&\bf {Human}&\bf {mean IoU}\\
		\hline
		fine-tune w/o   auxiliary loss       &   78.5&   72.2&   44.0&   65.3&   43.5&   17.4&   51.5&   1.2&   46.7\\
		%\hline
		fine-tune w     auxiliary loss       &   79.2&   72.5&\bf44.8&   64.6&   44.3&   17.0&   52.8&   3.4&   47.3\\
		%\hline
		fine-tune w+w/o auxiliary loss       &\bf79.4&\bf73.1&   43.7&\bf65.5&\bf45.3&\bf21.3&\bf55.8&\bf6.3&\bf48.8\\
		\hline
	\end{tabular}
}
\end{table*}

\subsection{Spatial-temporal Regularization for Semantic Segmentation}
For spatial-temporal regularization, we apply methods used in feature space optimization (FSO)~\cite{FSO}. As FSO process a block of images simultaneously, a block of 5 consecutive frames with a gap of 10 frames are extracted from the provided video files, and the test image is located at the center in each block. The gap between consecutive frames is not too big in order to get good flow extraction. 
It is better to have longer sequences to gain longer temporal regularization, but due to memory limitation, it is not possible to support more than 5 images in a 30G memory without sacrificing the image size.

The FSO process in each block requires several ingredients. Contour strength for each image is calculated according to ~\cite{edge}. The unary for each image is set as the softmax layer output from each fine-tuned network. Forward flows and backward flows are calculated according to ~\cite{LDOF1,LDOF2}. As the computation speed for optical flow at the original image scale is extremely low, the images to be processed are downsized by 8 times for both width and height, and the final flows at the original scale are calculated through bicubic interpolation and magnification. Then, points trajectories can be calculated according to ~\cite{track} with the forward and backward flows. Finally, a dense 3D CRF is applied after feature space optimization as described in ~\cite{FSO}.

\begin{figure*}%[thpb]
\centering
%\textcolor{red}{uncomment for image}
\includegraphics[width=1.0\textwidth]{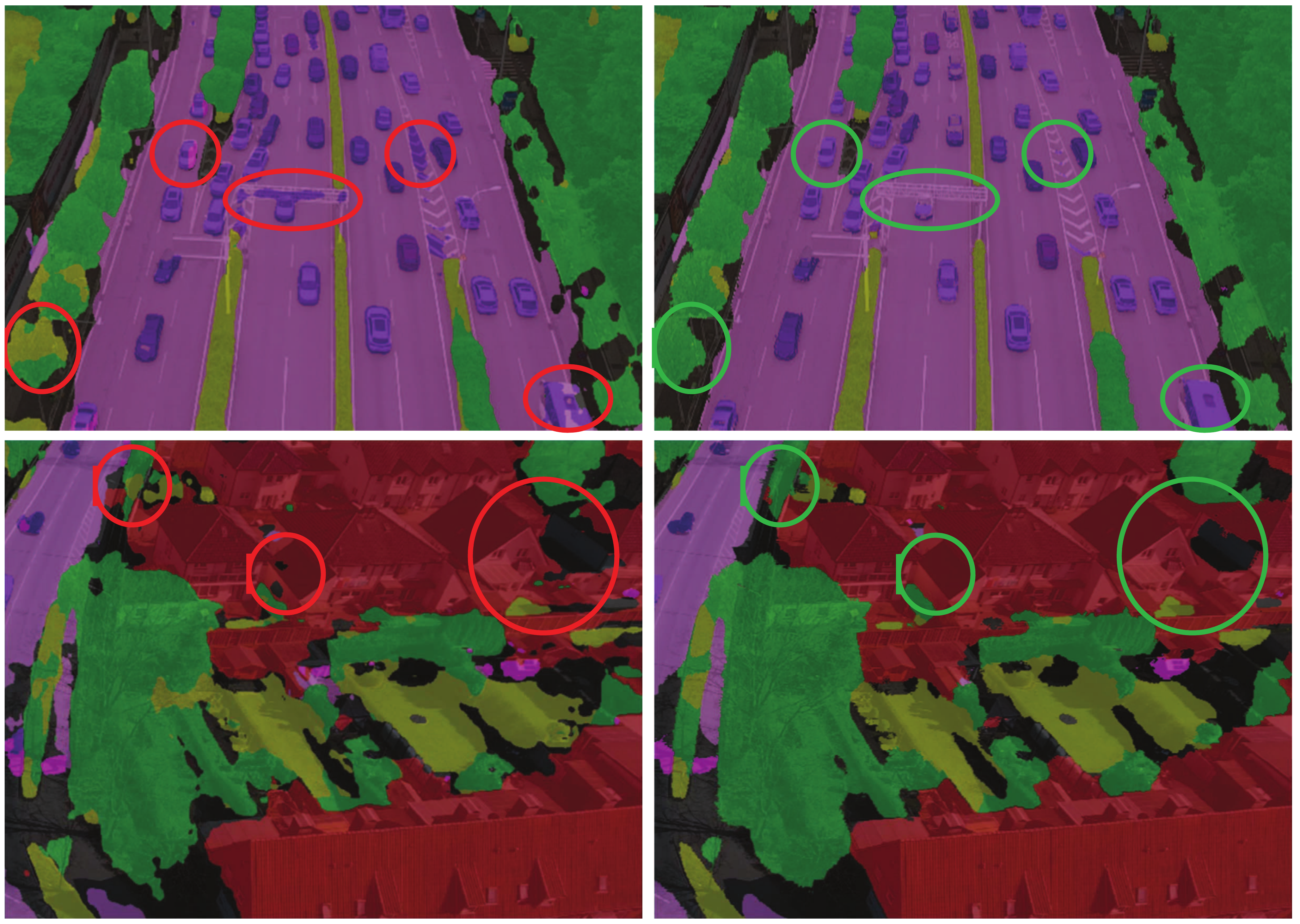}
\includegraphics[width=0.80\textwidth]{./legends}
\caption{Examples of spatial-temporal regularization for UAVid image semantic segmentation. The left column shows the prediction without FSO plus 3D CRF refinement. The right column shows the corresponding refined prediction with FSO plus 3D CRF refinement. The most obvious improvements are high-lighted with circles. The spatial-temporal regularization achieves a more coherent prediction for different objects.}
\label{fig_fso_eg}
\end{figure*}

The IoU scores for FSO post-processing with unaries from different fine-tuned networks are reported in Tab.~\ref{tb_IoU}. For each model, there is around $1\%$ IoU score improvement for each individual class except for human and moving car classes. FSO favors more for the class whose instance covers more image pixels. The IoU score improves less for the class with smaller instances like static car class, and it drops for moving car class and human class. The IoU score of the human class for  MS-Dilation net drops by a large margin, nearly $4\%$. An example of refinement is shown in Fig.~\ref{fig_fso_eg}.

\begin{figure*}%[thpb]
\centering
%\textcolor{red}{uncomment for image}
\includegraphics[width=1.0\textwidth]{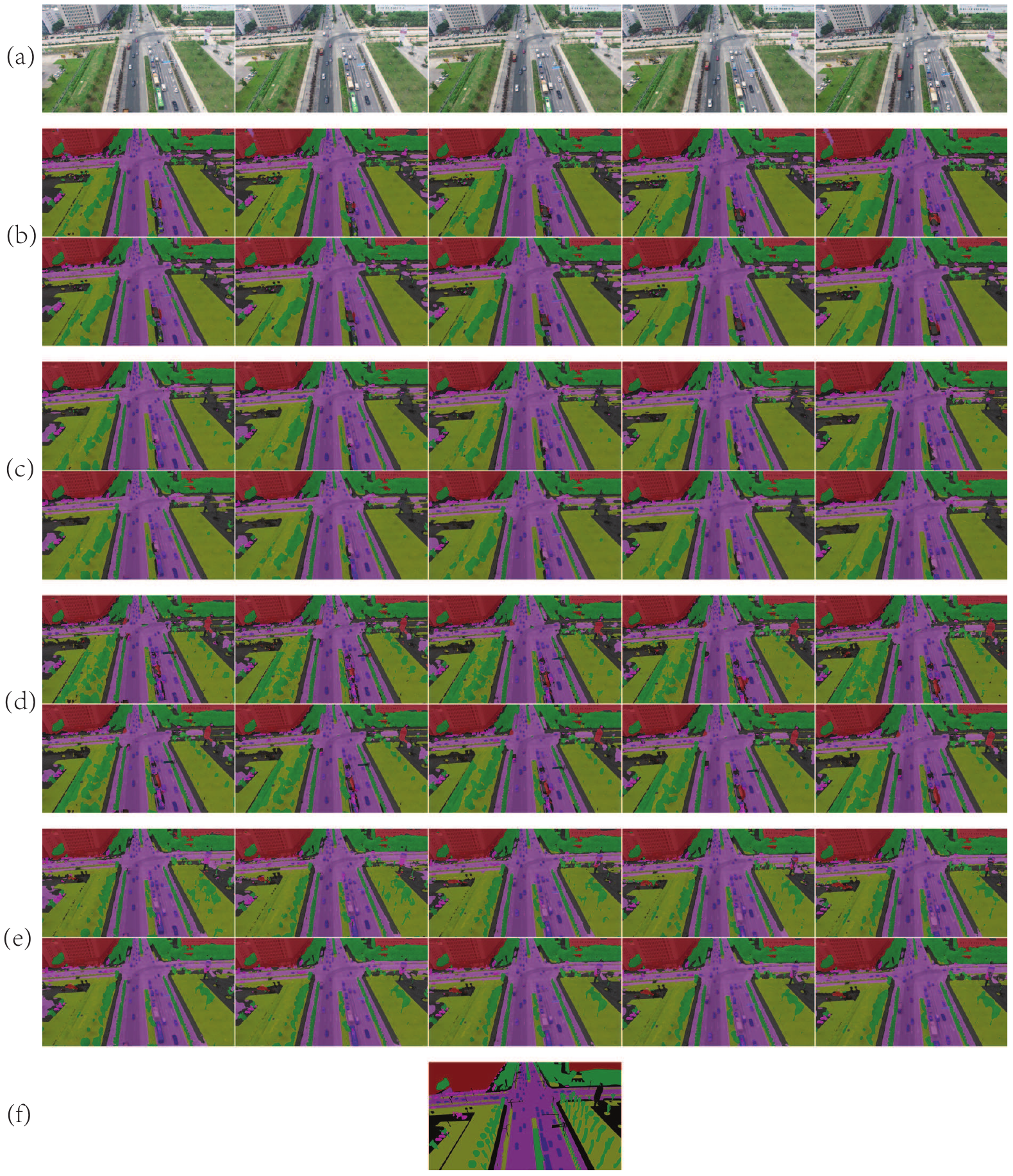}
\caption{\textbf{Example prediction for sequence images.} The 1st row block (a) presents the original images in sequential order from left to right. The last row block (f) presents the ground truth label for the test image located in the middle of the sequence. The 2nd, 3rd, 4th and 5th row blocks (b,c,d,e) present the prediction results of different models as it is in Tab.~\ref{tb_IoU}. Two rows from block (b) present prediction of FCN-8s+PRT and FCN-8s+PRT+FSO respectively. Two rows from block (c) present prediction of Dilation Net+PRT and Dilation Net+PRT+FSO respectively. Two rows from block (d) present prediction of U-Net+PRT and U-Net+PRT+FSO respectively. Two rows from block (e) present prediction of MS-Dilation Net+PRT and MS-Dilation Net+PRT+FSO respectively. PRT and FSO are defined the same as in Tab.~\ref{tb_IoU}.}
\label{fig_example_results_2}
\end{figure*}

%Example predictions for sequence images are shown in appendix.
In addition, qualitative prediction examples of different configurations across different time index are shown in Fig.\ref{fig_example_results_2}. Temporal consistency can be evaluated by viewing one row of the figure. Different model settings can be evaluated by viewing one column of the figure.

\section{Conclusions} ~\label{sec:conclusion}
In this paper, we have presented a new UAVid dataset to advance the development of semantic segmentation in urban street scenes from UAV images. Our dataset has brought out several challenges for the semantic segmentation task, including the large scale variation for different objects, the moving object recognition in the street scenes, and the temporal consistency across multiple frames. Eight classes for the semantic labeling task have been defined and labeled. The usability of our UAVid dataset has also been proved with several deep convolutional neural networks, among which the proposed Multi-Scale-Dilation net performs the best via multi-scale feature extraction. It has also been shown that pre-training the network and applying the spatial-temporal regularization are beneficial for the UAVid semantic labeling task. 
Although the UAVid dataset has some limitations in the size and the number of classes compared to the biggest dataset in the computer vision community, the UAVid dataset can already be used for benchmark purposes. In the future, we would like to further expand the dataset in size and the number of categories to make it more challenging and useful to advance the semantic segmentation research for the UAV imagery.

\section{Acknowledgments}
The work is partially funded by ISPRS Scientific Initiative project SVSB (PI: Michael Ying Yang, co-PI: Alper Yilmaz). The authors gratefully acknowledge the support. We also thank several graduate students from University of Twente and Wuhan University for their annotation effort.

\newpage
{\small
	\bibliographystyle{ieee_fullname}
	\bibliography{mybibfile}

\begin{thebibliography}{10}\itemsep=-1pt

\bibitem{tensorflow}
Mart{\'\i}n Abadi, Paul Barham, Jianmin Chen, Zhifeng Chen, Andy Davis, Jeffrey
  Dean, Matthieu Devin, Sanjay Ghemawat, Geoffrey Irving, Michael Isard, et~al.
\newblock Tensorflow: a system for large-scale machine learning.
\newblock In {\em OSDI}, volume~16, pages 265--283, 2016.

\bibitem{slic}
Radhakrishna Achanta, Appu Shaji, Kevin Smith, Aurelien Lucchi, Pascal Fua,
  Sabine S{\"u}sstrunk, et~al.
\newblock Slic superpixels compared to state-of-the-art superpixel methods.
\newblock {\em PAMI}, 34(11):2274--2282, 2012.

\bibitem{pyramid_images}
Edward~H Adelson, Charles~H Anderson, James~R Bergen, Peter~J Burt, and Joan~M
  Ogden.
\newblock Pyramid methods in image processing.
\newblock {\em RCA Engineer}, 29(6):33--41, 1984.

\bibitem{CamVid}
Gabriel~J. Brostow, Jamie Shotton, Julien Fauqueur, and Roberto Cipolla.
\newblock Segmentation and recognition using structure from motion point
  clouds.
\newblock In {\em ECCV}, pages 44--57, 2008.

\bibitem{LDOF1}
Thomas Brox, Andr{\'e}s Bruhn, Nils Papenberg, and Joachim Weickert.
\newblock High accuracy optical flow estimation based on a theory for warping.
\newblock In {\em ECCV}, pages 25--36, 2004.

\bibitem{LDOF2}
Thomas Brox and Jitendra Malik.
\newblock Large displacement optical flow: descriptor matching in variational
  motion estimation.
\newblock {\em PAMI}, 33(3):500--513, 2011.

\bibitem{pretrain_davis}
Sergi Caelles, Kevis-Kokitsi Maninis, Jordi Pont-Tuset, Laura Leal-Taix{\'e},
  Daniel Cremers, and Luc Van~Gool.
\newblock One-shot video object segmentation.
\newblock In {\em CVPR}, 2017.

\bibitem{coco_stuff}
Holger Caesar, Jasper Uijlings, and Vittorio Ferrari.
\newblock Coco-stuff: Thing and stuff classes in context.
\newblock In {\em CVPR}, June 2018.

\bibitem{zeebruges}
Manuel Campos-Taberner, Adriana Romero-Soriano, Carlo Gatta, Gustau
  Camps-Valls, Adrien Lagrange, Bertrand~Le Saux, Anne Beaupère, Alexandre
  Boulch, Adrien Chan-Hon-Tong, Stephane Herbin, Hicham Randrianarivo, Marin
  Ferecatu, Michal Shimoni, Gabriele Moser, and Devis Tuia.
\newblock Processing of extremely high-resolution lidar and rgb data: Outcome
  of the 2015 ieee grss data fusion contest part a: 2-d contest.
\newblock {\em IEEE Journal of Selected Topics in Applied Earth Observations
  and Remote Sensing}, 2016.

\bibitem{Algriculture}
Nived Chebrolu, Thomas L{\"a}be, and Cyrill Stachniss.
\newblock Robust long-term registration of uav images of crop fields for
  precision agriculture.
\newblock {\em IEEE Robotics and Automation Letters}, 3(4), 2018.

\bibitem{pretrain_cityscape}
Liang-Chieh Chen, Yukun Zhu, George Papandreou, Florian Schroff, and Hartwig
  Adam.
\newblock Encoder-decoder with atrous separable convolution for semantic image
  segmentation.
\newblock In {\em ECCV}, September 2018.

\bibitem{Cordts2016Cityscapes}
Marius Cordts, Mohamed Omran, Sebastian Ramos, Timo Rehfeld, Markus Enzweiler,
  Rodrigo Benenson, Uwe Franke, Stefan Roth, and Bernt Schiele.
\newblock The cityscapes dataset for semantic urban scene understanding.
\newblock In {\em CVPR}, 2016.

\bibitem{HoustonCampus}
Christian Debes, Andreas Merentitis, Roel Heremans, Jürgen Hahn, Nikolaos
  Frangiadakis, Tim van Kasteren, Wenzhi Liao, Rik Bellens, Aleksandra
  Pizurica, Sidharta Gautama, Wilfried Philips, Saurabh Prasad, Qian Du, and
  Fabio Pacifici.
\newblock Hyperspectral and lidar data fusion: Outcome of the 2013 grss data
  fusion contest.
\newblock {\em IEEE Journal of Selected Topics in Applied Earth Observations
  and Remote Sensing}, 7, 05 2014.

\bibitem{deepglobe}
I. Demir, K. Koperski, D. Lindenbaum, G. Pang, J. Huang, S. Basu, F. Hughes, D.
  Tuia, and R. Raska.
\newblock Deepglobe 2018: A challenge to parse the earth through satellite
  images.
\newblock In {\em CVPRW}, 2018.

\bibitem{edge}
Piotr Doll{\'a}r and C~Lawrence Zitnick.
\newblock Fast edge detection using structured forests.
\newblock {\em PAMI}, 37(8):1558--1570, 2015.

\bibitem{uav_det}
Dawei Du, Yuankai Qi, Hongyang Yu, Yifan Yang, Kaiwen Duan, Guorong Li, Weigang
  Zhang, Qingming Huang, and Qi Tian.
\newblock The unmanned aerial vehicle benchmark: object detection and tracking.
\newblock In {\em ECCV}, 2018.

\bibitem{PascalVOC}
M. Everingham, S.~M.~A. Eslami, L. Van~Gool, C.~K.~I. Williams, J. Winn, and A.
  Zisserman.
\newblock The pascal visual object classes challenge: A retrospective.
\newblock {\em IJCV}, 111(1):98--136, Jan. 2015.

\bibitem{kitti}
Andreas Geiger, Philip Lenz, Christoph Stiller, and Raquel Urtasun.
\newblock Vision meets robotics: The kitti dataset.
\newblock {\em International Journal of Robotics Research}, 32(11):1231--1237,
  2013.

\bibitem{aeroscapes}
Deva~Ramanan Ishan~Nigam, Chen~Huang.
\newblock Ensemble knowledge transfer for semantic segmentation.
\newblock In {\em WACV}, 2018.

\bibitem{HighwayDriving}
Byungju Kim, Junho Yim, and Junmo Kim.
\newblock Highway driving dataset for semantic video segmentation.
\newblock In {\em BMVC}, 2018.

\bibitem{FSO}
Abhijit Kundu, Vibhav Vineet, and Vladlen Koltun.
\newblock Feature space optimization for semantic video segmentation.
\newblock In {\em CVPR}, pages 3168--3175, 2016.

\bibitem{MSCoco}
Tsung-Yi Lin, Michael Maire, Serge Belongie, James Hays, Pietro Perona, Deva
  Ramanan, Piotr Doll{\'a}r, and C~Lawrence Zitnick.
\newblock Microsoft coco: Common objects in context.
\newblock In {\em ECCV}, pages 740--755, 2014.

\bibitem{pretrain_isprs}
Yongcheng Liu, Bin Fan, Lingfeng Wang, Jun Bai, Shiming Xiang, and Chunhong
  Pan.
\newblock Semantic labeling in very high resolution images via a self-cascaded
  convolutional neural network.
\newblock {\em ISPRS Journal of Photogrammetry and Remote Sensing}, 2018.

\bibitem{FCN8s}
Jonathan Long, Evan Shelhamer, and Trevor Darrell.
\newblock Fully convolutional networks for semantic segmentation.
\newblock In {\em CVPR}, June 2015.

\bibitem{smartfarm}
Philipp Lottes, Raghav Khanna, Johannes Pfeifer, Roland Siegwart, and Cyrill
  Stachniss.
\newblock Uav-based crop and weed classification for smart farming.
\newblock In {\em ICRA}, pages 3024--3031, 2017.

\bibitem{UAV_monitor_weed}
Andres Milioto, Philipp Lottes, and Cyrill Stachniss.
\newblock Real-time blob-wise sugar beets vs weeds classification for
  monitoring fields using convolutional neural networks.
\newblock {\em ISPRS Annals}, 4:41, 2017.

\bibitem{uav123}
Matthias Mueller, Neil Smith, and Bernard Ghanem.
\newblock A benchmark and simulator for uav tracking.
\newblock In Bastian Leibe, Jiri Matas, Nicu Sebe, and Max Welling, editors,
  {\em ECCV}, 2016.

\bibitem{UAV_Surveillance2}
Daniel Perez, Ivan Maza, Fernando Caballero, David Scarlatti, Enrique Casado,
  and Anibal Ollero.
\newblock A ground control station for a multi-uav surveillance system.
\newblock {\em Journal of Intelligent\&Robotic Systems}, 69(1-4):119--130,
  2013.

\bibitem{UAV_behavior}
Alexandre Robicquet, Amir Sadeghian, Alexandre Alahi, and Silvio Savarese.
\newblock Learning social etiquette: Human trajectory understanding in crowded
  scenes.
\newblock In {\em ECCV}, pages 549--565, 2016.

\bibitem{UNet}
Olaf Ronneberger, Philipp Fischer, and Thomas Brox.
\newblock U-net: Convolutional networks for biomedical image segmentation.
\newblock In {\em MICCAI}, pages 234--241, 2015.

\bibitem{ISPRSbenchmark14}
F. Rottensteiner, G. Sohn, M. Gerke, J.D. Wegner, U. Breitkopf, and J. Jung.
\newblock Results of the isprs benchmark on urban object detection and 3d
  building reconstruction.
\newblock {\em ISPRS journal of photogrammetry and remote sensing},
  93:256--271, 2014.

\bibitem{Daimler_Urban_Segmentation}
Timo Scharw{\"a}chter, Markus Enzweiler, Uwe Franke, and Stefan Roth.
\newblock Efficient multi-cue scene segmentation.
\newblock In {\em GCPR}, pages 435--445, 2013.

\bibitem{UAV_Surveillance1}
Eduard Semsch, Michal Jakob, Du{\v{s}}an Pavlicek, and Michal Pechoucek.
\newblock Autonomous uav surveillance in complex urban environments.
\newblock In {\em WI-IAT}, pages 82--85, 2009.

\bibitem{track}
Narayanan Sundaram, Thomas Brox, and Kurt Keutzer.
\newblock Dense point trajectories by gpu-accelerated large displacement
  optical flow.
\newblock In {\em ECCV}, pages 438--451, 2010.

\bibitem{GID}
Xin-Yi Tong, Gui-Song Xia, Qikai Lu, Huanfeng Shen, Shengyang Li, Shucheng You,
  and Liangpei Zhang.
\newblock Land-cover classification with high-resolution remote sensing images
  using transferable deep models.
\newblock {\em Remote Sensing of Environment}, 237:111322.

\bibitem{UAV_Monitoring}
Haitao Xiang and Lei Tian.
\newblock Development of a low-cost agricultural remote sensing system based on
  an autonomous unmanned aerial vehicle (uav).
\newblock {\em Biosystems Engineering}, 108(2):174--190, 2011.

\bibitem{dilationNet}
Fisher Yu and Vladlen Koltun.
\newblock Multi-scale context aggregation by dilated convolutions.
\newblock In {\em ICLR}, 2016.

\bibitem{deepdrive}
Fisher Yu, Wenqi Xian, Yingying Chen, Fangchen Liu, Mike Liao, Vashisht
  Madhavan, and Trevor Darrell.
\newblock Bdd100k: A diverse driving video database with scalable annotation
  tooling.
\newblock {\em arXiv preprint arXiv:1805.04687}, 2018.

\bibitem{pretrain_ade20k}
Hengshuang Zhao, Jianping Shi, Xiaojuan Qi, Xiaogang Wang, and Jiaya Jia.
\newblock Pyramid scene parsing network.
\newblock In {\em CVPR}, pages 2881--2890, 2017.

\bibitem{ADE20k}
Bolei Zhou, Hang Zhao, Xavier Puig, Sanja Fidler, Adela Barriuso, and Antonio
  Torralba.
\newblock Scene parsing through ade20k dataset.
\newblock In {\em CVPR}, volume~1, page~4, 2017.

\bibitem{visdrone}
Pengfei Zhu, Longyin Wen, Xiao Bian, Ling Haibin, and Qinghua Hu.
\newblock Vision meets drones: A challenge.
\newblock {\em arXiv preprint arXiv:1804.07437}, 2018.

\end{thebibliography}
}

\end{document}